\documentclass[letterpaper, 10 pt, conference]{ieeeconf}  

\IEEEoverridecommandlockouts                              

\usepackage{amsmath} 
\usepackage{graphics} 
\usepackage{times} 
\usepackage{amssymb}  
\usepackage[]{algorithm}
\usepackage{comment}
\usepackage{color}
\usepackage{multirow}

\usepackage{multicol}
\usepackage[bookmarks=true]{hyperref}
\usepackage{graphicx}
\usepackage[noend]{algpseudocode}
\usepackage{colortbl}
\usepackage{tabularx}
\usepackage{authblk}

\usepackage[dvipsnames]{xcolor}

\long\def\ignorethis#1{}

\newcommand{\etal}{{\em{et~al.}\ }}
\newcommand{\eg}{e.g.\ }
\newcommand{\ie}{i.e.\ }

\newcommand{\vc}[1]{\ensuremath{\mathbf{#1}}}

\newcommand{\pctab}{\hspace{0.2in}}

\title{\LARGE \bf
Data-Driven Approach to Simulating Realistic Human Joint Constraints
}

\author{Yifeng Jiang$^{1}$ and C. Karen Liu$^{2}$
\thanks{$^{1}${Yifeng Jiang is with School of Electrical and Computer Engineering, Georgia Institute of Technology, Atlanta, GA 30332, USA \tt\small yjiang340@gatech.edu}}%
\thanks{$^{2}${C. Karen Liu is with School of Interactive Computing,
Georgia Institute of Technology, Atlanta, GA 30332, USA \tt\small karenliu@cc.gatech.edu}}%
\thanks{Accompanying video: \url{https://youtu.be/wzkoE7wCbu0}} 
\thanks{Github repositories: \url{https://github.com/jyf588/Human-Joint-Constraints-Training} and \url{https://github.com/dartsim/dart/pull/1016}}
}

\begin{document}

\maketitle
\thispagestyle{empty}
\pagestyle{empty}

\begin{abstract}

Modeling realistic human joint limits is important for applications involving physical human-robot interaction. However, setting appropriate human joint limits is challenging because it is pose-dependent: the range of joint motion varies depending on the positions of other bones. The paper introduces a new technique to accurately simulate human joint limits in physics simulation. We propose to learn an implicit equation to represent the boundary of valid human joint configurations from real human data. The function in the implicit equation is represented by a fully connected neural network whose gradients can be efficiently computed via back-propagation. Using gradients, we can efficiently enforce realistic human joint limits through constraint forces in a physics engine or as constraints in an optimization problem.
\end{abstract}

\section{Introduction}
As robots move from industrial applications to providing personal assistance alongside people, the emergence of collaborative robotics demands more generalizable yet tractable methodologies to model human movements and behaviors. The robot control policy that considers human state based on such methodologies has the potential to provide more effective assistance when applied in the real world. For a collaborative scenario that involves physical contacts and forceful interactions between the robot and the human, accurately predicting human movements is paramount to not only the functionality of the robot but also the safety of the human.


Modeling human behaviors in response to robot actions highly depends on the task of interest, the functionality of the robot, and the environment features and constraints. However, one common concern faced by most collaborative scenarios involving physical interactions is the awareness of the range of human motion due to physical joint limits. If the human joint limits are modeled too conservatively, the robot might miss out many effective strategies to assist people. On the other hand, overly relaxed human joint limits might lead to a robot control policy unsafe to humans. Setting appropriate human joint limits is challenging because not only that different joints have different ranges of angles, biomechanics literature \cite{engin1986statistical}\cite{wang1998three}\cite{hatze1997three} suggests that the range of angle varies depending on the positions of other joints (inter-joint dependency) or other degrees-of-freedom in the same joint (intra-joint dependency). For example, the range of flexion of our elbow depending on whether it is in front of or behind the body.


The paper introduces a new technique to enhance a general-purpose physics engine, such as Bullet \cite{Bullet}, MuJoCo \cite{Mujoco}, or DART \cite{DART}, to accurately simulate human as a dynamic system. As joint limits are one of the most important kinematic constraints that give rise to the unique characteristics of human movements, our technique formulates realistic joint limits from real human data and enforces them through constraint forces in a dynamic system. Our work is built on the comprehensive study on the range of human motion conducted by Akhter and Black \cite{akhter2015pose}, who captured human motion that includes an extensive variety of stretching poses performed by trained athletes. Using the dataset, they developed a procedure to determine whether a human configuration is within the valid range. This validity procedure involves discrete operations such as “if statements” and table lookups. While it is sufficient for determining the validity of a given pose, the lack of analytical representation and the non-differentiable nature of this procedure prevents it from being incorporated into the process of physics simulation.

We propose to learn an implicit equation, $C(\vc{q}) = 0$, to represent the boundary of valid human joint configurations, where $C(\vc{q})$ is an analytical and differentiable function. We utilize the validity procedure developed by Akhter and Black to provide unlimited labeled data for learning $C(\vc{q})$ represented as a fully connected neural network. The main benefit of a neural network representation is that the gradient of the function can be easily computed via back-propagation. Using gradients, we can efficiently enforce realistic human joint limits through constraint forces in a physics engine. For any model-free policy learning method, human joint limits will be enforced as part of the "black-box" physics simulation, similar to the way contact constraints are handled. In addition to dynamic applications, we can also utilize the gradients of the neural network to solve inverse kinematics (IK) problems where solutions are confined in the set of valid human poses.

Our method is general, computationally efficient, and easy to implement. The learned joint-limit constraint can achieve $95\%$ accuracy. We evaluate the joint-limit constraint as a dynamic constraint in a physics simulation and as a kinematic constraint in a pose optimization.

\section{Related Work}


Human joint ranges vary significantly from facet joints being able to merely rotate several degrees \cite{kozanek2009range} to shoulder joints capable of near 180 degree movements in all three degrees-of-freedom \cite{snell2007clinical}. The intra-joint and inter-joint dependencies further complicate the range of each degree-of-freedom \cite{engin1986statistical}\cite{wang1998three}\cite{hatze1997three}. For example, Hatze \cite{hatze1997three} showed the inter-joint coupling between the elbow range and the shoulder orientation. A model of viscoelastic torques of the elbow was also proposed, but the method requires to estimate numerous subject-specific parameters from repeated experiments, making it difficult to be applied as a general joint-limit constraint.


An accurate and tractable computational model of human joint limits has many applications in computer vision and graphics. Enforcing joint limits has been found useful to disambiguate 3D poses estimated from 2D images \cite{chen2013data}\cite{rehg2003ambiguities}
\cite{sminchisescu2003estimating}\cite{demirdjian2003enforcing}. However, most methods in this area assume a fixed range for each degree-of-freedom of the joint, overly simplifying the range of human motion to a few box constraints. Human pose distribution learned from motion capture data has also been heavily used in motion reconstruction in computer animation \cite{grochow2004style}\cite{zhang2014leveraging}\cite{chai2007constraint}\cite{min2009interactive}\cite{brand2000style}. For example, Grochow \etal \cite{grochow2004style} used this distribution as a prior to solve constrained IK problems in order to generate human-like natural poses. Nevertheless, none of these methods aim to precisely define the boundary of the range of human motion. In contrast, our work enforces the learned joint limits in physics simulation, which might enable applications in robotics, such as the design of collaborative robots.

A few more realistic joint-limit models have been proposed recently. Brau and Jiang \cite{brau2016bayesian} incorporated both joint orientation and self-occlusion as prior distributions to estimate 3D poses, because joint limits are caused by anatomical constraints and by the physical presence of other body parts. However, it is not clear how one can enforce the validity of a pose from the priors. Herda \etal \cite{herda2005hierarchical} modeled arm joint limits as implicit hypersurfaces learned from motion capture data. The range of elbow is modeled as an array of implicit equations, indexed by discretized shoulder orientations. While this method might suffice kinematic applications, it does not directly apply to dynamic applications using physics simulation. Akhter and Black \cite{akhter2015pose} also utilized human data to develop a procedure that determines whether a full-body pose is within the valid range of human motion. Their discontinuous model can be used as an inequality constraint, but cannot utilize efficient gradient-based methods to solve the optimization. Built upon Akhter and Black's work, our analytical and differentiable model can be incorporated widely in any constraint solving or optimization problems that demand accurate gradients.

\section{Method}
We take a data-driven approach to learn a hypersurface represented by an implicit equation, $C(\vc{q}) = 0$, where \vc{q} indicates the joint configuration of the agent in generalized coordinates. The hypersurface represents the boundary of the range of motion a human can achieve. If $\vc{q}$ is a valid configuration, $C(\vc{q}) \geq 0$. Otherwise, $C(\vc{q})$ is negative.

Once a differentiable, analytical function, $C(\vc{q})$ is learned, we can utilize its gradient in a number of robotic applications. For example, we can create a dynamic constraint to enforce the learned joint limits in a physics simulator. We can also enforce joint limits as constraints in planning or trajectory optimization problems.

\subsection{Learning joint limits}
\label{sec:method-learning}
The function $C(\vc{q})$ is represented by a fully connected neural network, which maps a joint configuration $\vc{q}$ to a scalar which indicates the validity of $\vc{q}$. We utilize the method developed by Akhter and Black \cite{akhter2015pose} to generate unlimited training data. Using an extensive motion capture dataset of human poses, Akhter and Black introduced a procedure, $isValid(\vc{p}): \mathbb{R}^{3\times (N+1)} \mapsto \{0,1\}^N$, which maps an array of 3D joint positions, $\vc{p}$ from $N$ bone segments, to an array of binary numbers indicating the validity of the orientation of each bone segment. Their algorithm only evaluates the validity of the arms without hands, the legs, and the head. It also assumes that the validity of each limb is independent of other limbs or the head. As such, we define two functions $C_{arm}(\vc{q}): \mathbb{R}^4 \mapsto [-0.5,0.5]$ and $C_{leg}(\vc{q}): \mathbb{R}^6 \mapsto [-0.5,0.5]$ for the four limbs of the agent. The input of $C_{arm}(\vc{q})$ consists of a shoulder with 3 degrees-of-freedom (3-DOF) and a 1-DOF elbow, while the input of $C_{leg}(\vc{q})$ considers a 3-DOF hip, a 1-DOF knee, and a 2-DOF ankle. Evaluating the validity of the neck is not presented in this paper but is a trivial extension. 

The procedure for generating training samples is shown in Algorithm 1. A training sample is a pair of joint configuration and its label. For each limb with $N$ bone segments, we first initialize $N+1$ buffers, $D_0, \cdots, D_N$, for storing the training samples categorized by the validity of the joint configuration (Line 1). According to Akhter and Black, if a bone segment is invalid, all the offspring bones are considered invalid. This treatment can be justified by the fact that the validity of a child segment might depend on its parent segment but not vice versa. As such, there are only $N$ categories of invalid configurations, which are stored in $D_1$ to $D_N$. The samples with a valid configuration are stored in $D_0$. For example, the types of validity for the arm include valid upper arm and valid lower arm (stored in $D_0$), valid upper arm and invalid lower arm (stored in $D_1$), and invalid upper arm (stored in $D_2$).

To generate each training sample, the algorithm samples a vector, $\vc{q}$, from the joint configuration space (Line 3), computes the Cartesian positions of the joints from $\vc{q}$ via forward kinematics (Line 4), and evaluates the validity of the joint positions by calling the function $isValid(\vc{p})$ (Line 5). If all the bone segments are valid, the sample is labeled with value $1$ and stored in $D_0$ (Line 6-7). Otherwise, we label the sample $0$ and store it in the buffer corresponding to the most significant bit (MSB) of the inverted binary vector returned by $isValid(\vc{p})$ (Line 8-10). Note that the joint configuration is represented by the sine and cosine of each joint angle. This is a common practice to represent cyclic variables in the input space of a neural network to improve learning efficiency.

\setlength{\textfloatsep}{0pt}
\begin{algorithm}[t]
\caption{Generating Training Data}\label{alg:learning_alg}
\begin{algorithmic}[1]
\State Initialize $N+1$ buffers: $D_0, \cdots, D_N$
\While{$k < K$}
\State Uniformly sample $\vc{q}$ from joint configuration space\;
\State \vc{p} = ForwardKinematics($\vc{q}$) \;
\State \vc{b} = isValid(\vc{p}) \;
\If{all($\vc{b} == 1$)}
\State Store $([\sin(\vc{q}), \cos(\vc{q})], 1)$ in $D_0$ \;
\Else
\State $i$ = MSB($\scriptsize{\sim} \vc{b}$) $+1$ \;
\State Store $([\sin(\vc{q}), \cos(\vc{q})], 0)$ in $D_i$ \;
\EndIf
\State $k = \min(|D_0|, \cdots, |D_N|)$ \;
\EndWhile
\State Keep first $K$ samples in each buffer \;
\\
\Return{$D_0, \cdots, D_N$}
\end{algorithmic}
\end{algorithm}

As one might expect, the uniform sampling in the joint configuration space results in unbalanced distribution across categories $D$'s. For example, only $3\%$ of the uniformly sampled leg samples fall in $D_0$ and $9\%$ in $D_1$. On the other hand, $53\%$ of samples fall in $D_3$ and $35\%$ in $D_2$. The unbalanced samples will significantly bias the learning results. We use rejection sampling to balance the number of samples in each category. In practice, we sample uniformly until every buffer has at least $K$ samples and reject the extra samples in the buffers that have more than $K$ samples.

\setlength{\textfloatsep}{20pt plus 2pt minus 4pt}

We train two separate fully connected feed-forward neural networks for $C_{arm}(\vc{q})$ and for $C_{leg}(\vc{q})$. Each network has three hidden layers with 128 hidden units per layer. All activations are tanh except for the sigmoid function at the final layer. The output of the neural network is subtracted by 0.5 such that $C(\vc{q})$ ranges between $-0.5$ and $0.5$. Since the procedure to generate training data is completely automatic, the number of training samples can be as large as necessary. In our experiences, we find that $400,000$ samples in total is sufficient to reach $95\%$ accuracy on the testing set after $30$ epochs. The training of each neural network takes only a few minutes on a Core i5 CPU without GPU support.

Despite that the learned functions, $C_{arm}(\vc{q})$ and $C_{leg}(\vc{q})$, can well represent $isValid(\vc{p})$, there is a potential issue due to the ambiguity caused by the transformation from the joint configuration to the Cartesian joint positions (\ie $\vc{q} \mapsto \vc{p}$). For example, the joint configuration of shoulder rotation $180^\circ$ and elbow bending $-90^\circ$ results in the same 3D joint positions as those from the configuration of shoulder rotation $0^\circ$ and elbow bending $90^\circ$. $isValid(\vc{p})$ deems both configurations valid, but bending the elbow backward is clearly not achievable by humans. Fortunately, this ambiguity can be easily resolved if we also enforce standard box constraints on the joints during usage. These box constraints are set wider than the range of motion and are only used to exclude clearly impossible poses. For example, in our experiments we loosely define the box constraint of the rotation DOF in the shoulder to be $[-60^\circ, 120^\circ]$ and of the DOF in the elbow to be $[0^\circ, 180^\circ]$. We do not set box constraints on the other two DOFs in the shoulder. A valid joint configuration must satisfy both the learned joint-limit constraint $C(\vc{q}) > 0$ and the default box constraints. This also implies that sampling outside the box constraints during training is unnecessary.

\subsection{Enforcing joint limits as dynamic constraints}
Given the learned function, $C(\vc{q}): \mathbb{R}^n \mapsto [-0.5, 0.5]$, the implicit equation $C(\vc{q}) = 0$ represents a hyperpsurface in the joint configuration space where one side to the surface ($C(\vc{q}) < 0$) is infeasible. In physics simulation, this is analogous to a contact constraint that prevents two objects inter-penetrating each other. To maintain such dynamic constraints during simulation, we must apply constraint impulses to the simulated bodies when they are exactly on the hypersurface (\ie $C(\vc{q}) = 0$). The constraint impulses can be computed in different ways, one of which formulates a Linear Complementarity Problem (LCP) to solve all the active constraints at the current time step. In this paper, we incorporate our learned joint-limit constraints in the LCP framework, but they can be applied to other constraint handling methods, such as the optimization-based approach proposed by Todorov \cite{todorov2014convex}.

The governing equation in LCP is the relationship between the rate of constraint violation ($v$) and the constraint impulse that stops the violation ($f$). In an implicit-stepping formulation, we define $v$ as the rate of joint-limit violation in the next time step:
$$
v = \dot{C}(\vc{q}) = (\frac{\partial C}{\partial \vc{q}}) \dot{\vc{q}},
$$
where $\dot{\vc{q}}$ is the joint velocity at the next time step. The Jacobian, $\frac{\partial C}{\partial \vc{q}}$, can be efficiently calculated via back propagation of the trained network. 

When the constraint is active, \ie $C(\vc{q})=0$, LCP solves for $v$ and $f$ simultaneously such that the following three conditions are met:
\begin{enumerate}
\item $v \geq 0$: The unilateral joint-limit constraint must be satisfied at the next time step.
\item $f \geq 0$: The constraint impulse must only push in the direction that prevents constraint violation.
\item $v \cdot f=0$: Either $v$ or $f$ must be zero. If $v>0$, the joint limit will no longer be violated and thus there should be no constraint impulse ($f = 0$). If $f > 0$, the joint-limit constraint must continue to be active ($v = 0$).
\end{enumerate}	

Once the LCP is solved, we can compute the corresponding joint torques by:
$$
\boldsymbol{\tau} = (\frac{\partial C}{\partial \vc{q}})^T f,
$$
where $\boldsymbol{\tau}$ is a vector in $\mathbb{R}^n$. By applying $\boldsymbol{\tau}$ to the corresponding joints, the pose-dependent joint limits are enforced via dynamics. Note that the joint-limit constraints are solved together with other constraints (\eg contact constraints) and the dynamic equations of motion, such that the next state of the agent is dynamically valid within the range of motion.

\subsection{Enforcing joint limits in inverse kinematics (IK)}

Satisfying joint-limit constraints is important in kinematic trajectory planning. A typical planning algorithm uses IK to compute a reference trajectory which is then tracked by feedback controllers. Incorporating the joint-limit constraint in IK planning prevents the agent from hitting joint limits during execution:
\begin{align}
 \min_{\vc{q}} \;\;G(\vc{q}) = \| \vc{F}(\vc{q}) - \bar{\vc{p}} \|^2  \nonumber \\
\nonumber  \mathrm{subject\;} \mathrm{to\;\;} C(\vc{q}) \geq 0,
\end{align}
where $\vc{F}(\vc{q})$ is the forward kinematics routine that computes the Cartesian location of a body point from a joint configuration $\vc{q}$ and $\bar{\vc{p}}$ is the target location for the body point.

This non-convex constrained optimization can be solved by various optimization techniques, such as Sequential Quadratic Programming (SQP) or interior point methods. Here we simply enforce the constraint as a penalty in the objective function when the inequality is violated:
\begin{align}
 \min_{\vc{q}} \;\;G(\vc{q}) = \|\vc{F}(\vc{q}) - \bar{\vc{p}} \|^2 + w \| C(\vc{q}) \|^2,  \nonumber \\
\mathrm{where\;\;\;} w = \begin{cases} 
0, & C(\vc{q}) > 0\\
0.2, & C(\vc{q}) \leq 0
\end{cases}. \nonumber
\end{align}

In practice, we make the inequality constraint slightly tighter, $C(\vc{q}) > 0.02$, to strictly enforce the joint limits.

\section{Evaluation}
We first validated that the learned networks are an adequate differentiable substitute for $isValid()$ by reporting the accuracy on the test datasets. We then evaluated the joint-limit constraints represented by the networks in two applications: physics simulation and inverse kinematics. 

\subsection{Accuracy of classification}
\label{sec:eval-learning}
\begin{figure}[b]
\centering
  \includegraphics[width=0.42\textwidth]{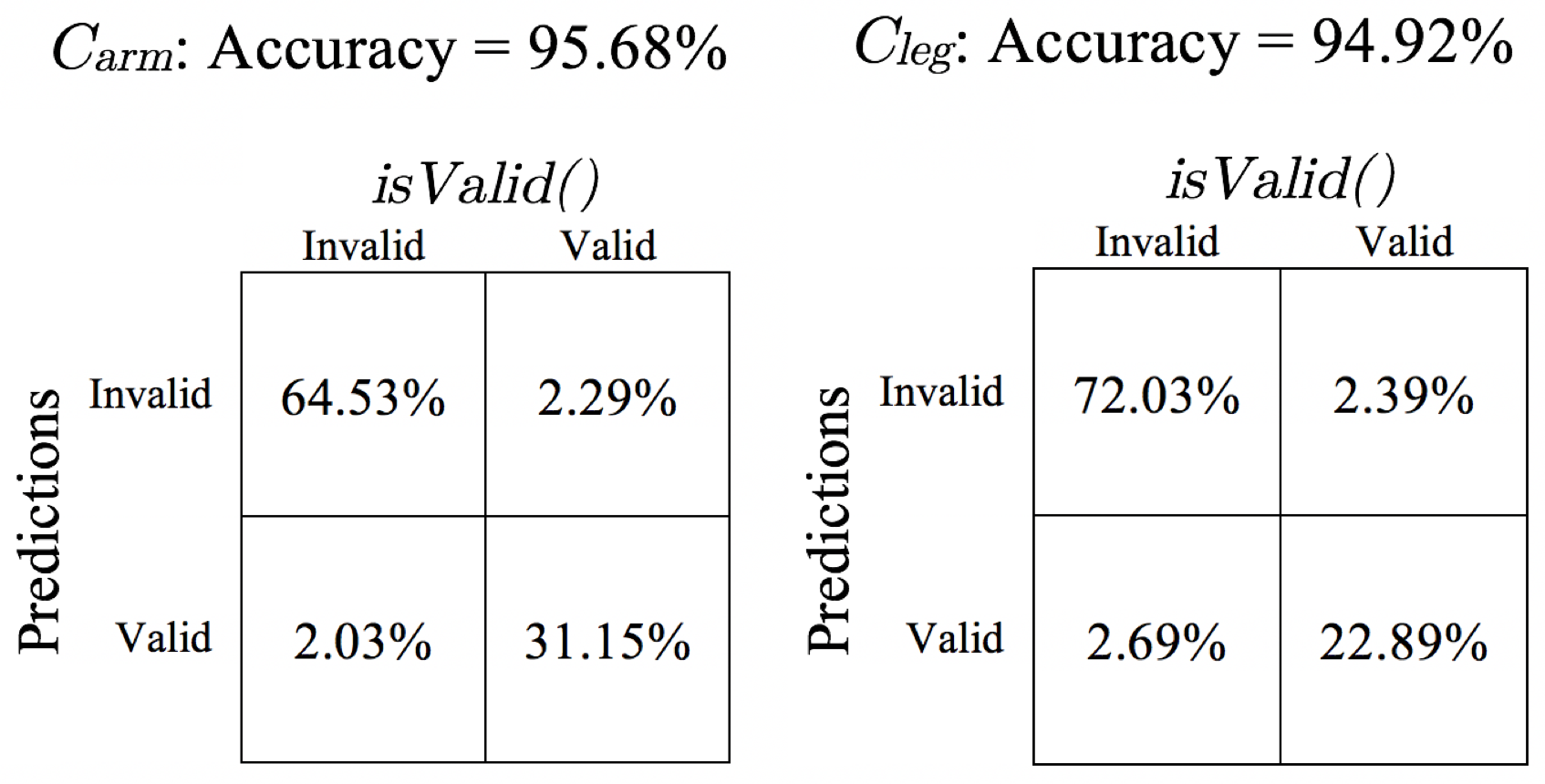}
  \caption{Confusion matrices of $C_{arm}(\vc{q})$ (Left) and $C_{leg}(\vc{q})$ (Right).}
  \label{fig:confusion_matrix}
\end{figure}

\begin{figure}[tbh]
\centering
  \includegraphics[width=0.42\textwidth]{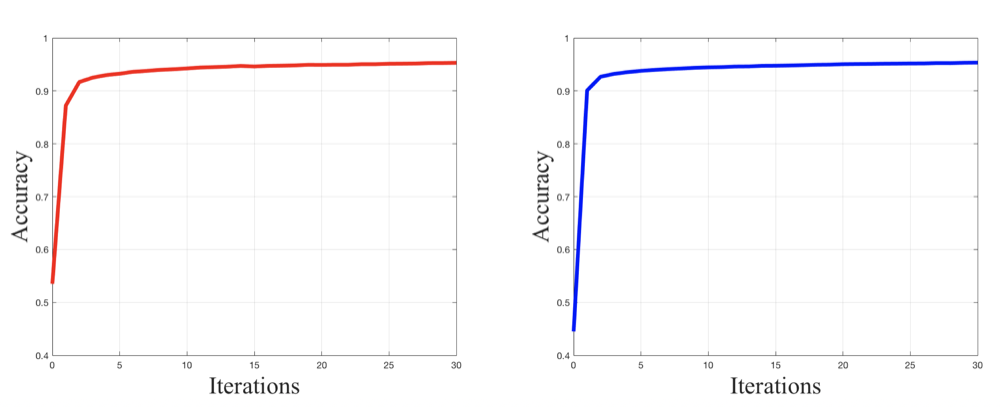}
  \caption{Learning curves of $C_{arm}(\vc{q})$ (Left) and $C_{leg}(\vc{q})$ (Right). }
  \label{fig:learning-curve}
\end{figure}

After training the neural networks using TensorFlow \cite{tensorflow2015-whitepaper}, we tested each neural network on a set of random joint configurations sampled from the same distribution as the training data. The confusion matrices for $C_{arm}(\vc{q})$ and $C_{leg}(\vc{q})$ are given in Fig. \ref{fig:confusion_matrix}. Rejection sampling reduces bias in the trained classifiers by balancing the false negative and false positive ratio. Fig. \ref{fig:learning-curve} shows the learning curves of $C_{arm}(\vc{q})$ and $C_{leg}(\vc{q})$.


\subsection{Physics simulation}
We used the open source physics engine DART \cite{DART}, which provides APIs to implement user-defined constraints without modifying the core code that formulates and solves LCP. For feedforward and back-propagation operations on a neural network, we incorporated the light-weight, C++ library tiny-dnn \cite{tiny-dnn}, with the trained weights imported from TensorFlow. To clearly visualize the effect of joint-limit constraints, the following experiments only simulate one limb with the torso fixed in place.

\begin{figure}[tbh]
\centering
  \includegraphics[width=0.45\textwidth]{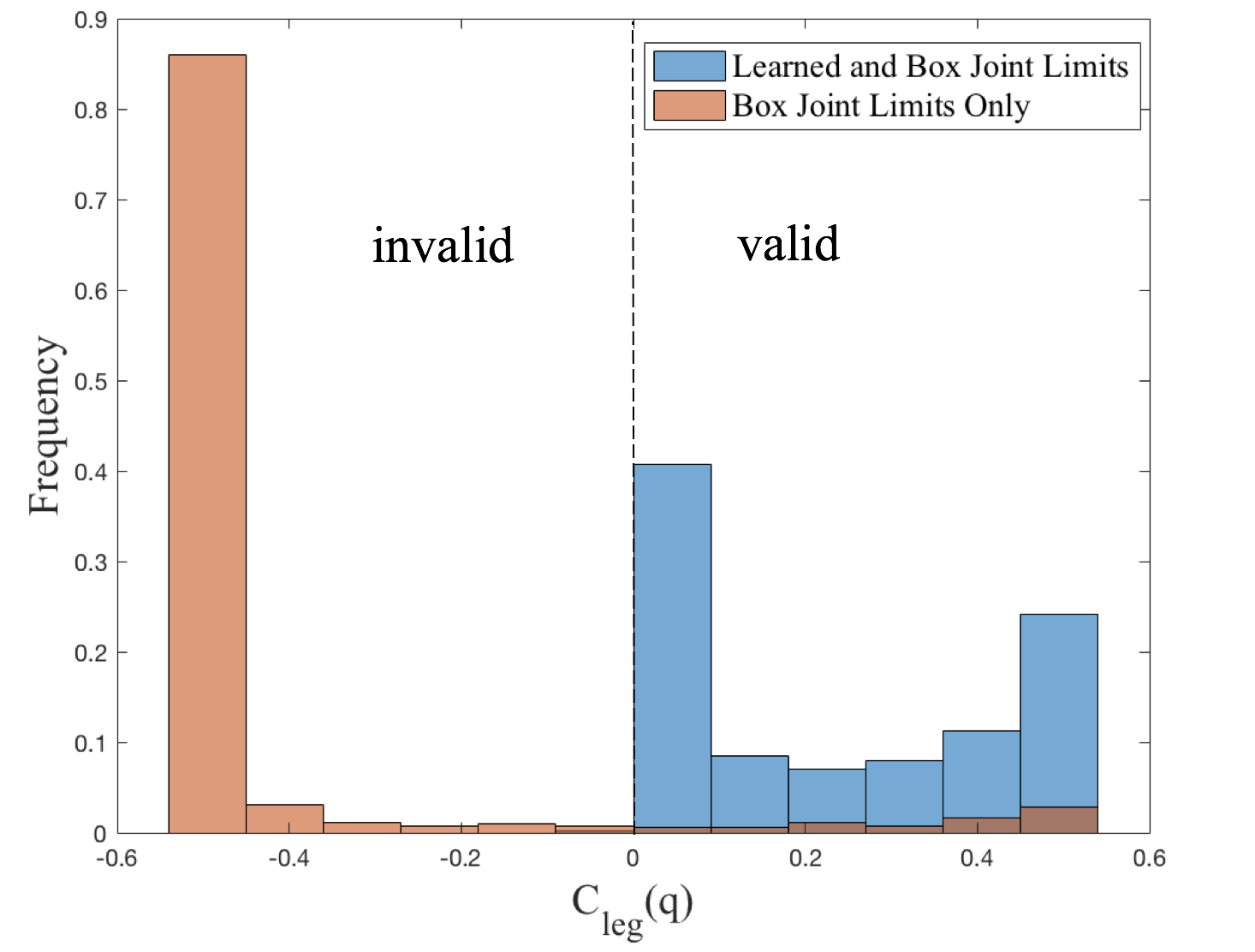}
  \caption{Histogram of joint-limit violation. The orange bars show the simulated joint configurations with only box constraints on joints enforced, which results in a large portion of invalid joint configurations. With our learned joint-limit constraint enforced (blue bars), all the tested configurations stay in the valid region of the constraint.}
  \label{fig:leg-FD-stats}
\end{figure}

\paragraph{Satisfaction of joint limits} To evaluate how precisely the joint-limit constraints are enforced in the simulation, we applied random joint torques to the agent and reported the statistics of joint-limit violation. With over $40,000$ joint configurations sampled along the simulated trajectory, Fig. \ref{fig:leg-FD-stats} shows the histogram of $C_{leg}(\vc{q})$, indicating the distribution of joint-limit violation. It is evident that the random joint torques frequently drive the agent to invalid joint configurations without the proposed joint-limit constraints. We use trained $C(\vc{q})$ as ground truth because it has been established in \ref{sec:eval-learning} that $C(\vc{q})$ is an accurate approximator of $isValid()$.

\begin{figure}[tbh]
\centering
  \includegraphics[width=0.42\textwidth]{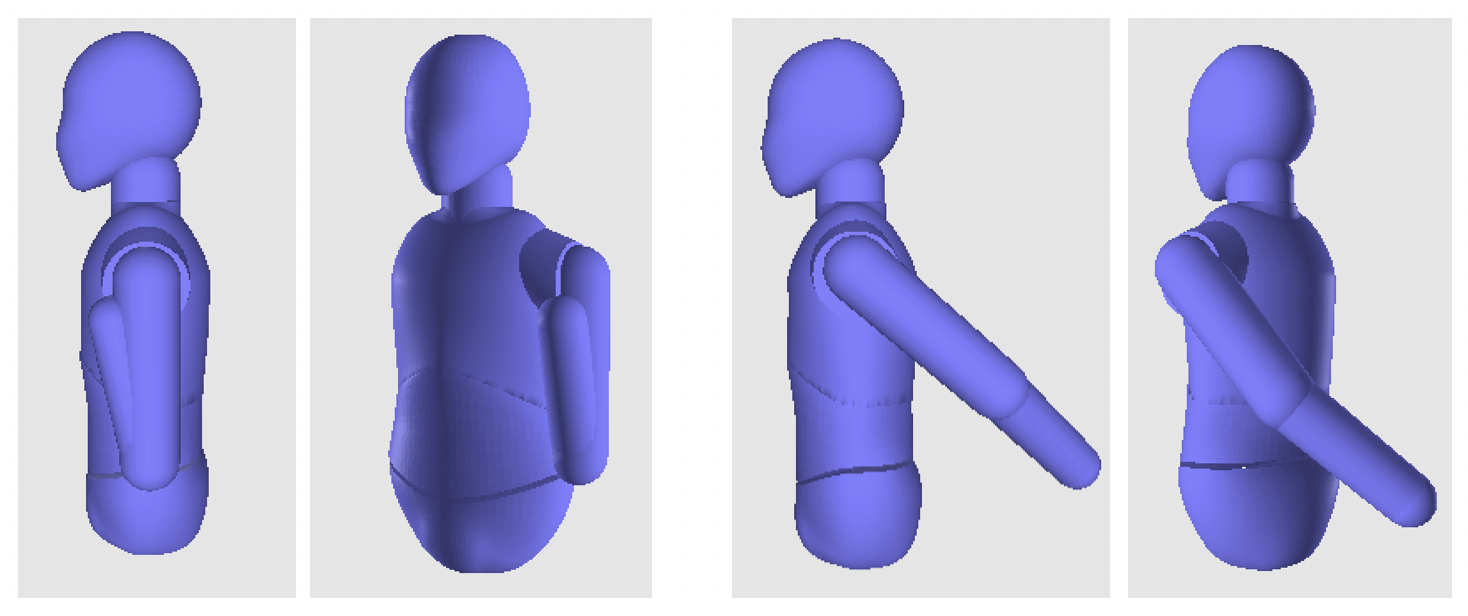}
  \caption{Pose-dependent elbow flexion limits. Left: When the arm is beside the torso, the elbow can flex up to $180^{\circ}$. Right: When the arm is behind the torso, the elbow can only flex up to $35^{\circ}$.}
  \label{fig:elbow-range}
\end{figure}

\begin{figure}[tbh]
\centering
  \includegraphics[width=0.42\textwidth]{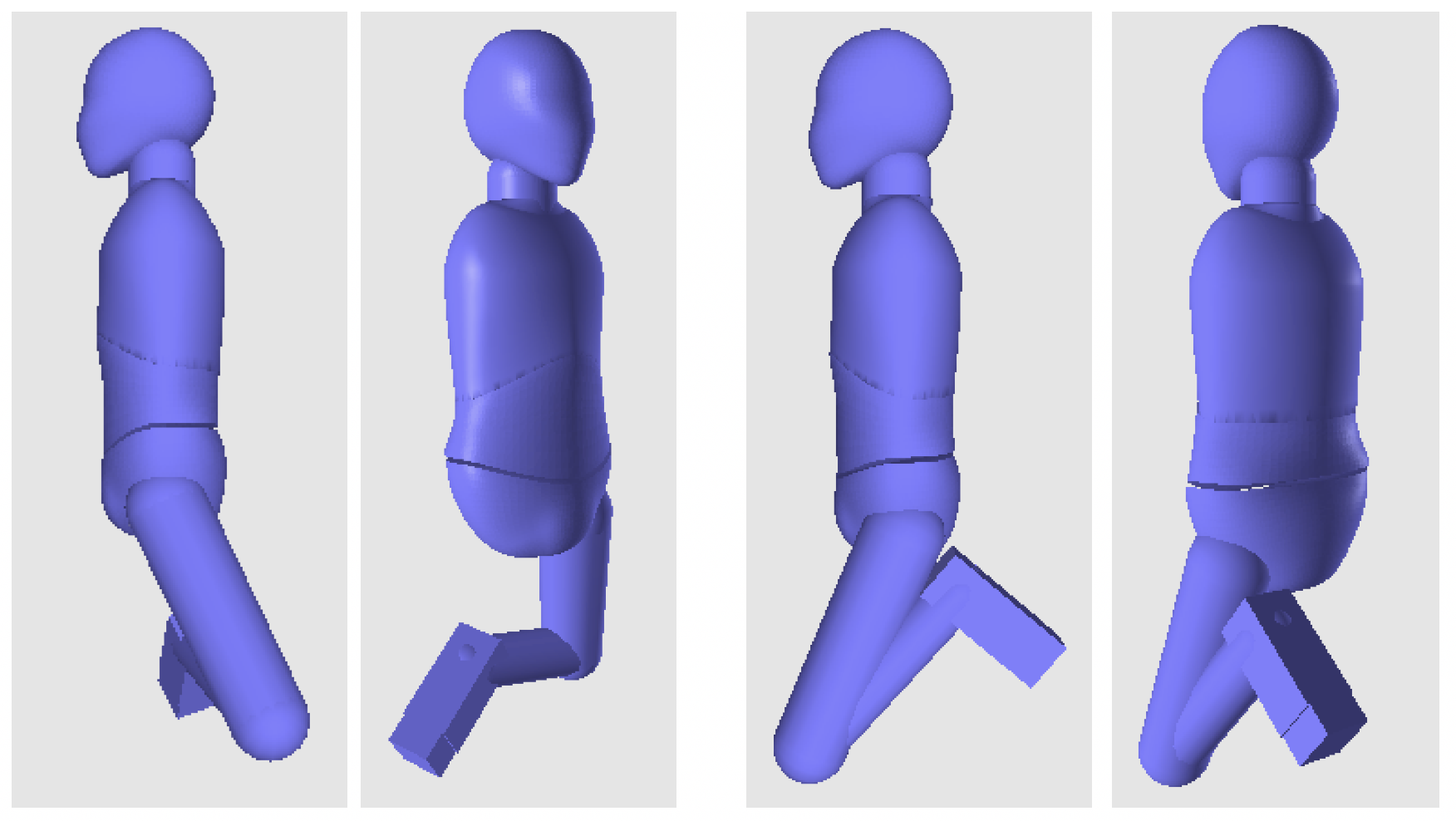}
  \caption{Pose-dependent knee flexion limits. Left: When the hip is abducting, the knee can only flex up to $125^{\circ}$. Right: When the hip is flexing, the knee can flex up to $165^{\circ}$.}
  \label{fig:knee-range}
\end{figure}

\paragraph{Pose-dependent joint range} Akhter and Black \cite{akhter2015pose} used the term "pose-conditioned" to emphasize that some joint ranges depend on the configuration of their parent joints. To test whether our joint-limit constraints exhibit the same dependency, we simulated elbow flexion with different fixed upper arm positions. Fig. \ref{fig:elbow-range} shows the moments when the elbow limit is reached. When the upper arm is beside the torso, the elbow can flex much more than when the upper arm is behind the torso. Similarly, Fig. \ref{fig:knee-range} shows that the knee has a wider range of motion when the hip is flexing as opposed to abducting.

\begin{figure}[tbh]
\centering
  \includegraphics[width=0.41\textwidth]{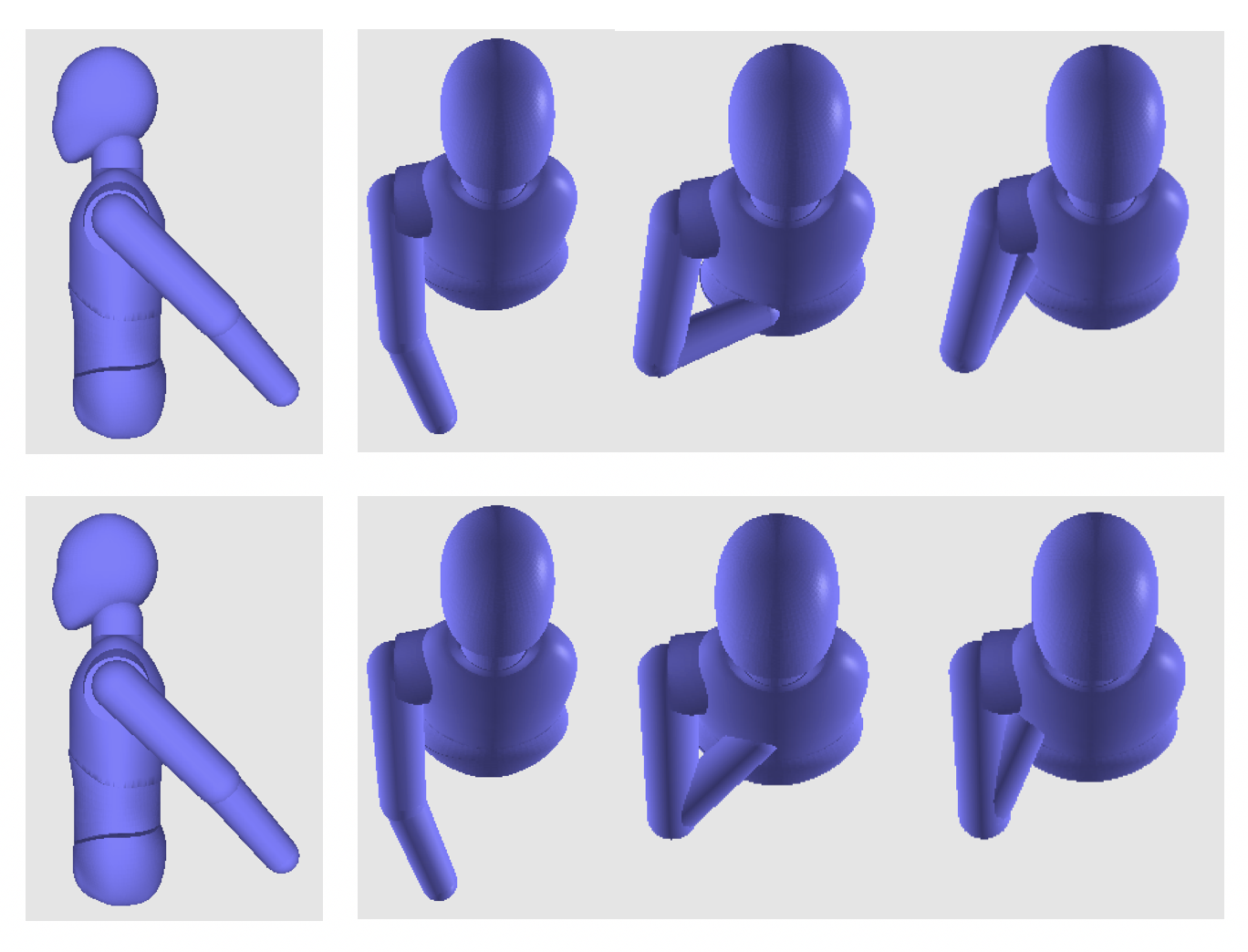}
  \caption{Shoulder unlocking. Torque is applied on the elbow when the arm is behind the back. Top: With the proposed joint-limit constraint, as the elbow continues to flex, the upper arm starts to abduct and rotate back towards the front, adjusting its position to allow more elbow flexion. Bottom: Without the joint-limit constraint, the upper arm stays at its initial position while the elbow flexes beyond the range of human motion.}
  \label{fig:elbow-around-invalid}
\end{figure}

\paragraph{Emergence of realistic motion} In this experiment, the agent applied torque on the elbow when the arm is behind the back. Unlike the previous example shown in Fig. \ref{fig:elbow-range} Right, we unlocked the upper arm and left it completely passive. As the elbow continues to flex, we observe that the upper arm starts to abduct and rotate back towards the front, adjusting its position to allow more elbow flexion (Fig. \ref{fig:elbow-around-invalid} Top). In contrast, without our joint-limit constraint, the upper arm stays in the same position while the elbow flexes beyond the range of human motion (Fig. \ref{fig:elbow-around-invalid} Bottom).

\subsection{Inverse kinematics}
In this set of experiments, the agent is commanded to reach a 3D location by solving an inverse kinematics problem. We formulated the IK problem as an optimization and used gradient descent to solve it.

\begin{figure}[tbh]
\centering
  \includegraphics[width=0.45\textwidth]{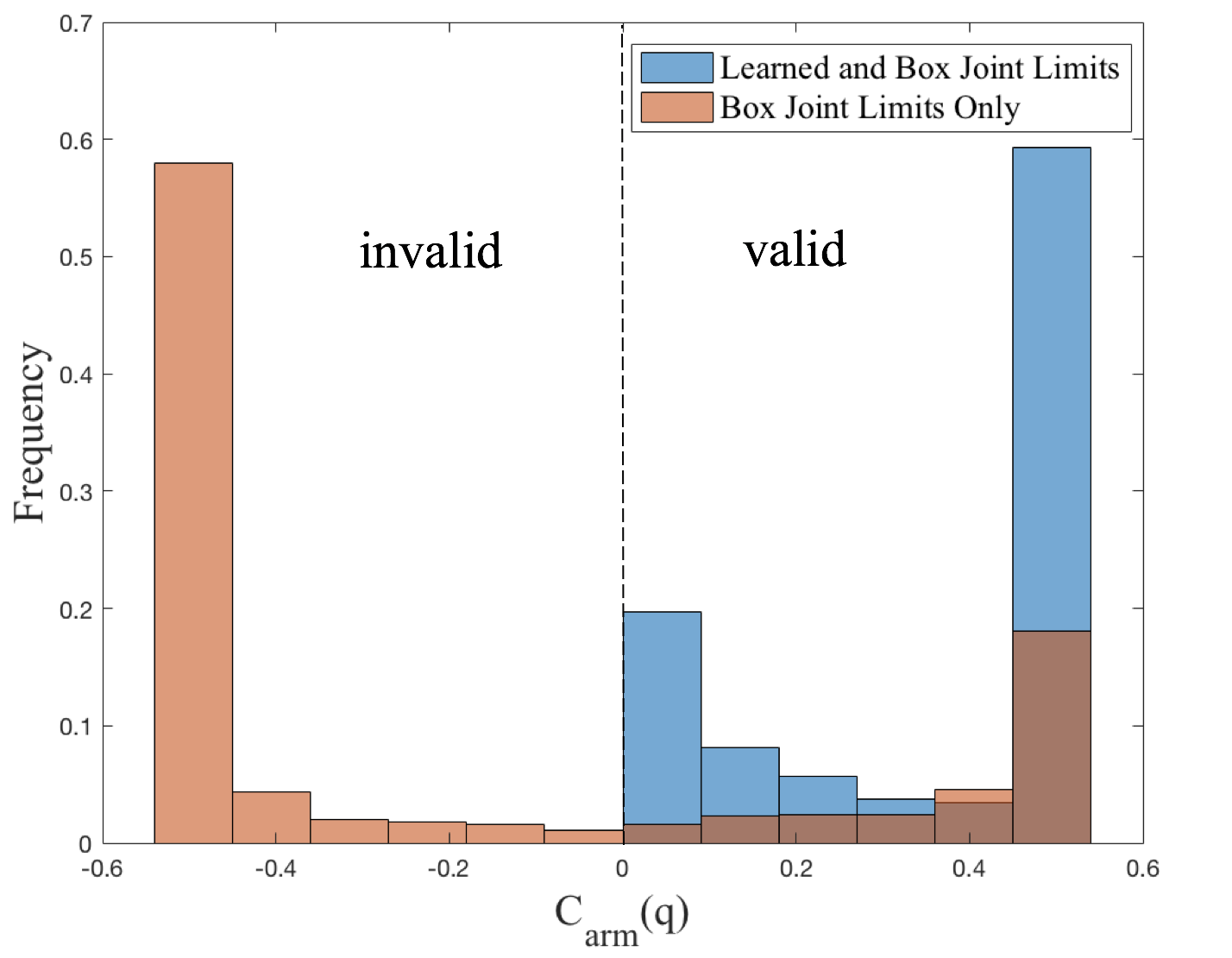}
  \caption{Histogram of joint-limit violation. The orange bars show the IK solutions without enforcing our joint-limit constraint. With the constraint enforced (blue bars), all the IK solutions are valid joint configurations.}
  \label{fig:arm-IK-stats}
\end{figure}

\paragraph{Satisfaction of joint limits} We randomly generated a set of 3D target locations and solved IK for each location sequentially. The joint configurations generated by solving these IK problems were then evaluated by the constraint functions $C(\vc{q})$. Fig. \ref{fig:arm-IK-stats} shows the histogram of $C_{arm}(\vc{q})$. The joint configurations strictly stay in the valid region when the optimization incorporates the learned joint-limit constraints as penalty. In contrast, most of the joint configurations are invalid when only box joint-limit constraints are enforced.

\begin{figure}[tbh]
\centering
  \includegraphics[width=0.41\textwidth]{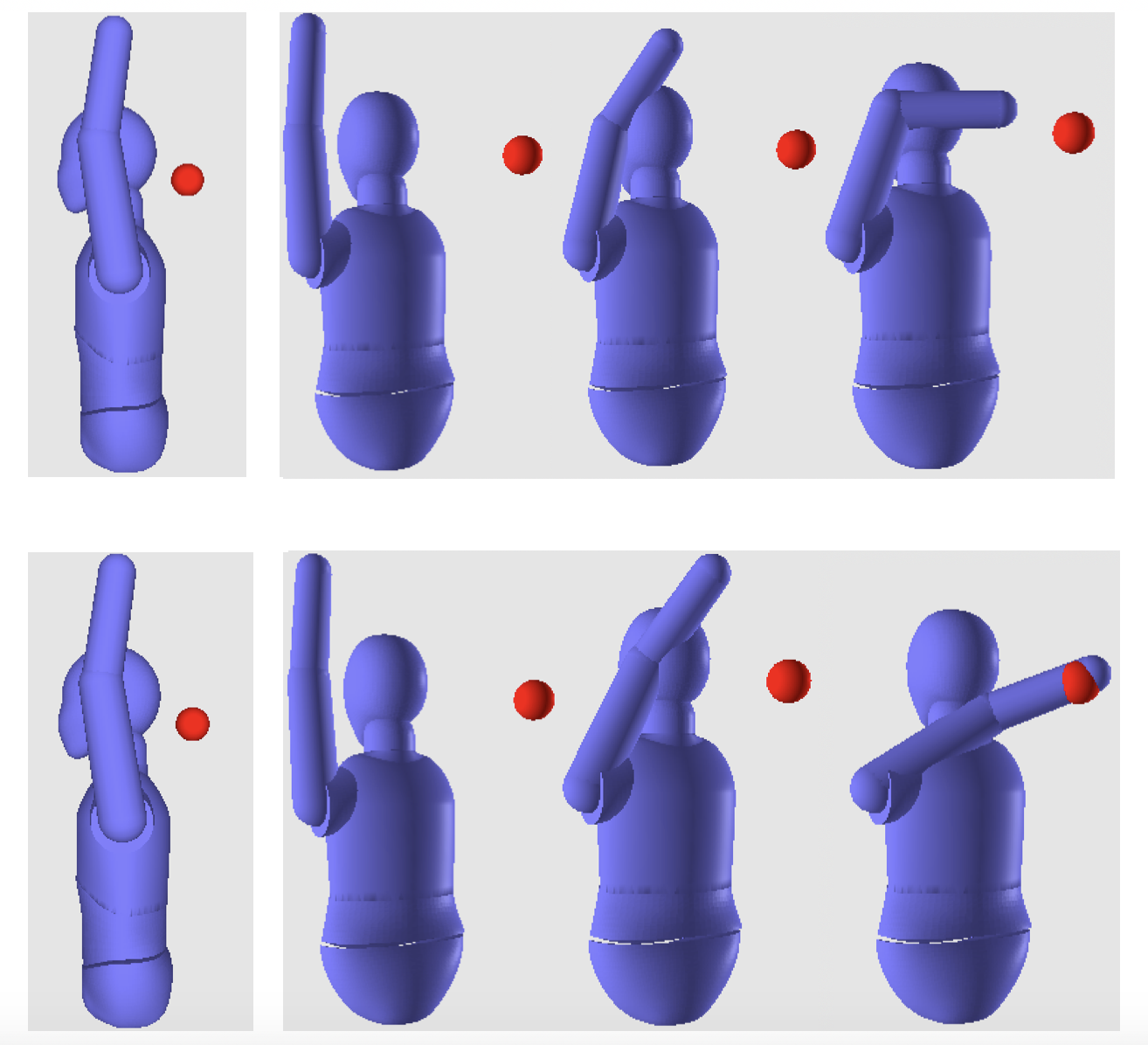}
  \caption{Unreachable target within the range of human motion. Top: With the learned joint-limit constraint, the IK solution that satisfies the joint limits "correctly" fails to reach the target. Bottom: With only box limits, the IK solution reaches the target by an unrealistic pose.}
   \label{fig:impossible-reach}
\end{figure}

\begin{figure}[tbh]
\centering
  \includegraphics[width=0.40\textwidth]{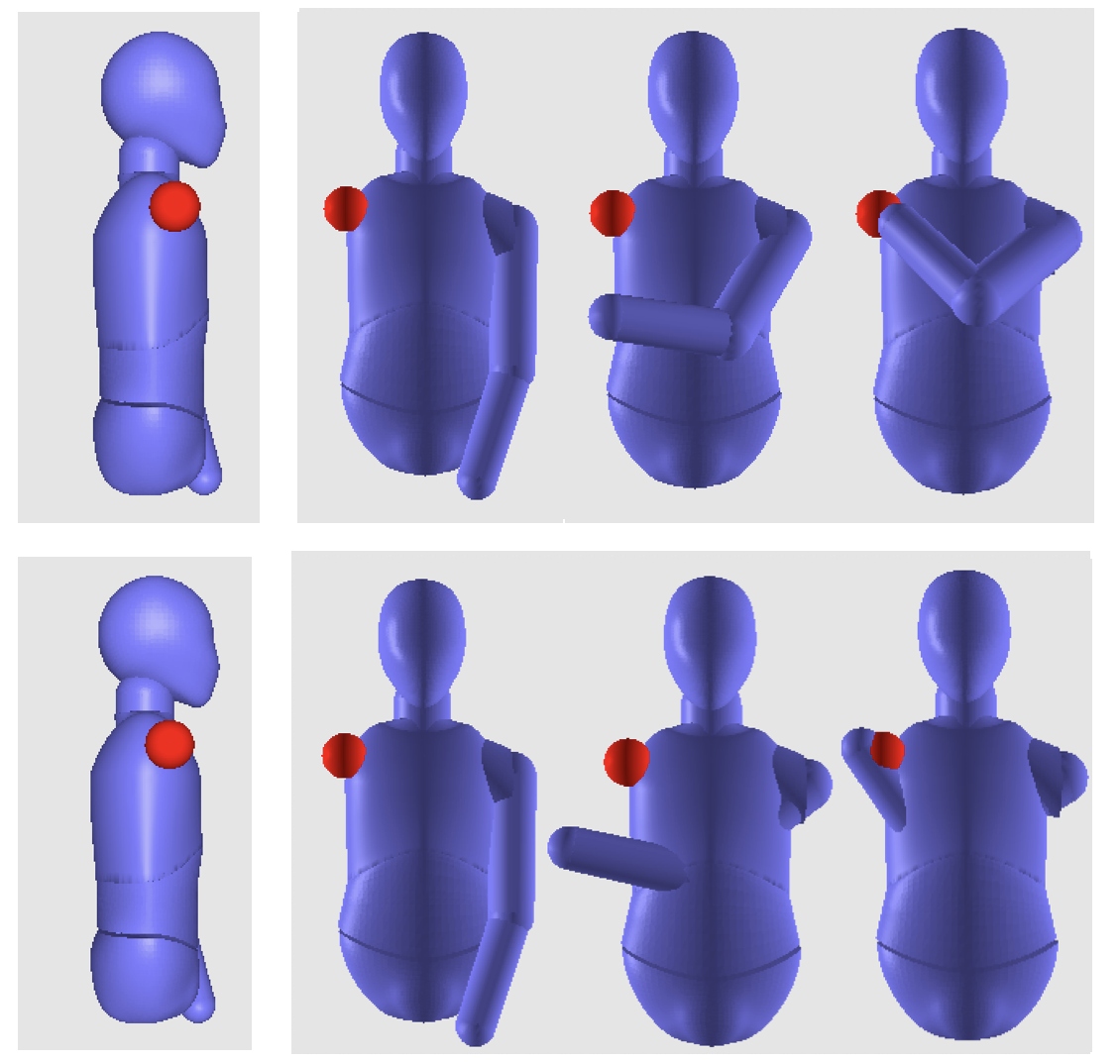}
  \caption{Self-collision. Top: Without activating collision detection and handling, our joint-limit constraint alone can make IK solution collision-free. Bottom: The IK solution with box limits only exhibits self-penetration.}
  \label{fig:self-collision}
\end{figure}

\paragraph{Emergence of realistic motion} Fig. \ref{fig:impossible-reach} demonstrates the situation where the target is unreachable within the range of human motion. Without our joint-limit constraint, the IK solution reaches the target by an unrealistic pose (Bottom). The IK solution that satisfies our joint limits "correctly" fails to reach the target (Top). In addition, we observe that our joint-limit constraints can mitigate the issue of self-collision because many of the self-penetrating poses are not presented in the database $isValid()$ learns from. As a result, our joint-limit constraints also classify these poses as invalid. Fig. \ref{fig:self-collision} shows that, without activating collision detection and handling, our joint-limit constraints alone can make IK solution collision-free.

\section{Conclusions}
This paper has proposed a general, computationally efficient, and easy to implement method to model accurate human joint limits. We have learned a joint-limit constraint function from real world data and represented it as a neural network. The differentiability of the function allows us to compute the gradient which is required for enforcing the joint limits using constraint forces in a physics simulation. In addition, we have shown that the joint-limit constraint can be incorporated into pose optimization problems and solved by gradient-based optimization methods.

Our method can be further improved. The trained network is specific to a particular joint configuration space with specific joint types, axis orders, and the rest pose. However, the proposed algorithm is general to learn the function for different joint configuration spaces. For example, one future direction is to use quaternion to represent ball joints to avoid the Gimbal lock. For another possible improvement, our learned function can evaluate the feasibility of a pose, but not the comfort level associated with it. Learning a function to address the preference of the human poses can be a possible future direction.

\addtolength{\textheight}{-12cm}   





\bibliographystyle{ieeetr}
\bibliography{reference}

\end{document}